\documentclass{article}

\usepackage{natbib}

\usepackage[a4paper, total={6in, 8in}]{geometry}

\usepackage[utf8]{inputenc}
\usepackage[T1]{fontenc}
\usepackage{hyperref}
\usepackage{url}
\usepackage{booktabs}
\usepackage{amsfonts}
\usepackage{nicefrac}
\usepackage{microtype}
\usepackage{xcolor}

\usepackage{wrapfig}
\usepackage{amsmath}

\usepackage{soul}

\DeclareMathAlphabet{\mathbbold}{U}{bbold}{m}{n}

\newcommand{\indep}{\perp\!\!\!\!\perp} 
\usepackage{mathtools}
\usepackage{amsmath}
\usepackage{algorithm}
\usepackage{algpseudocode}

\title{Tractable Hierarchical Control of Autoregressive Language Models}

\author{Max Scribner \\ m.v.scribner@sms.ed.ac.uk\and
   Antonio Vergari \\ avergari@ed.ac.uk\and
   Vaishak Belle \\ vbelle@ed.ac.uk \and
   University of Edinburgh}

\begin{document}

\maketitle

\begin{abstract}
Constraining the generation of autoregressive large language models (LLMs) is an important component of integrating language models into formal systems.
In the generation of code and data for tasks like program synthesis, ensuring that language models produce syntactically valid output is a prerequisite for processing such output. These languages (such as SQL or JSON) are often designed as $LR(k)$ context-free grammars.
By distilling the LLM to a tractable probabilistic model, its autoregressive generation can be steered and masked to incorporate the probability of satisfying logical constraints, ensuring high quality output that is guaranteed to be valid. 
This paper demonstrates that the satisfaction of any $LR(k)$ grammar of finite duration can be calculated in polynomial time, an improvement over the exponential time of applying previous methods to such grammars. This result enables efficient constraint and steering of LLM generation towards output that better satisfies formal syntactic constraints.
\end{abstract}

\section{Introduction}
\par
The output of large language models (LLMs) often needs to be parsed and converted into a structured form. 
In code generation \citep{10.1145/3747588}, data generation \citep{li2024large}, and structured natural language processing tasks \citep{geng-etal-2023-grammar}, large language models must generate sequences which satisfy a predetermined formal language to be processed for downstream tasks. Many of these languages are designed to be $LR(k)$ languages, i.e.,
languages which can be parsed using a shift-reduce parser \citep{10.5555/1177220}. The class of automata that classify $LR(k)$ languages is \textit{deterministic pushdown automata} (DPDAs) \citep{KNUTH1965607}. 
For this reason, improving the autoregressive generation of text that can be guaranteed to satisfy an arbitrary DPDA is important to the future of using the output of these models in formal systems. 

Even when prompted to satisfy simple hierarchical constraints, LLMs can fail, especially as the constraints stray out of their training distribution. Across four open LLMs (Figure 1), we found the models consistently failed to generate text satisfying a simple context-free grammar constraint, generating a short well-nested sequence of square brackets and angle brackets, called a Dyck-2 language (discussed in \cite{ebrahimi-etal-2020-self}). We found that as the set of parentheses changes from the more canonical $()[]$ set to less attested parentheses in the literature on Dyck languages such as $<>$ and $\{\}$, all models perform worse at generating well-typed and well-nested sequences. Despite the canonical Dyck-2 language showing up in these models' corpora, they fail to generate sequences satisfying the non-canonical but conceptually equivalent constraint requested, as illustrated in Figure 1. These models' inability to reliably generate text satisfying such constraints indicates the necessity of constrained and steered generation more broadly in tasks that require LLMs to satisfy more complex languages with such nested structures.

\begin{figure}[h]
\includegraphics[width=12cm]{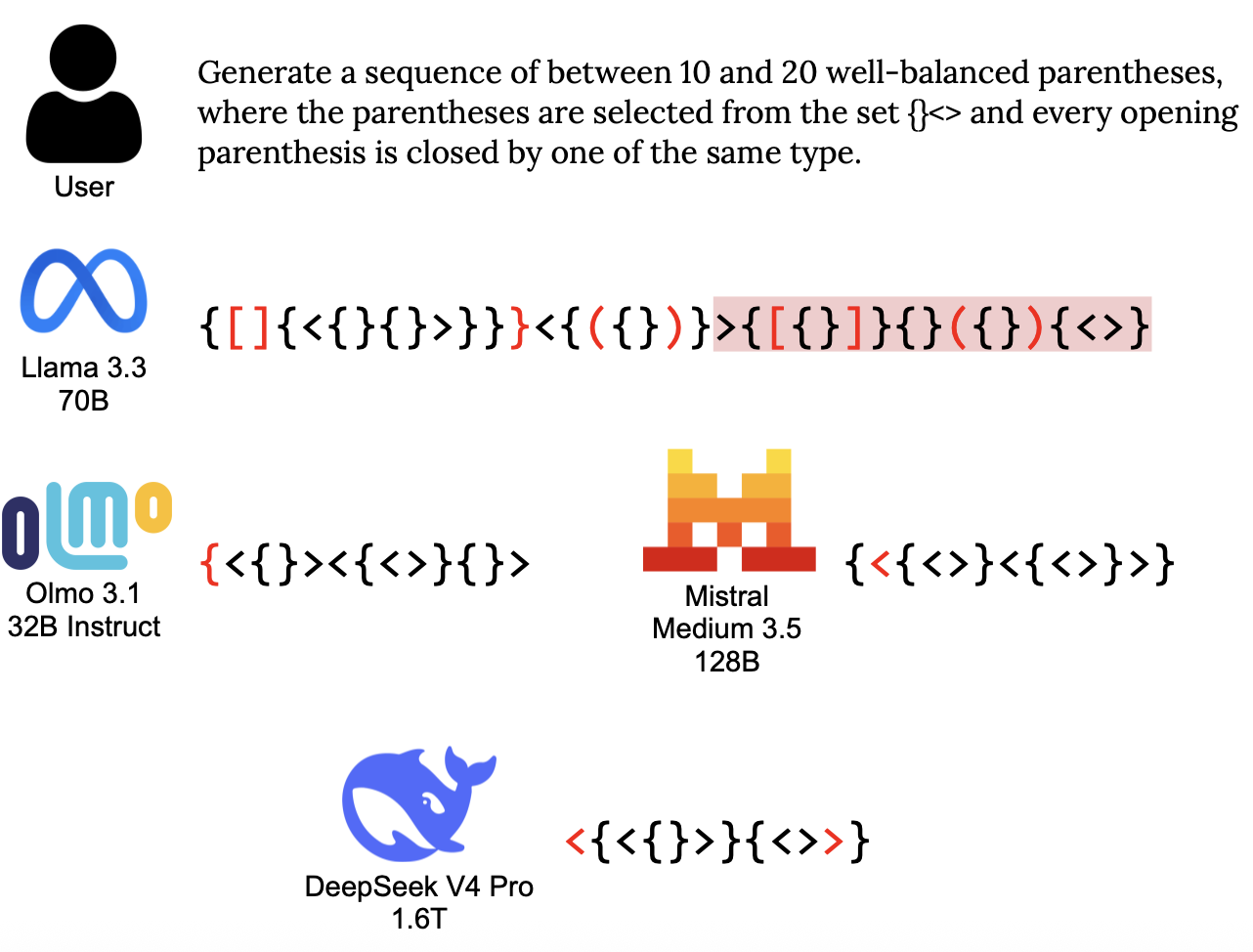}
\centering
\caption{\textbf{LLMs fail to satisfy simple context-free constraints}. Llama 3.3 (70 billion parameters), Olmo 3.1 (32 billion parameters), Mistral Medium 3.5 (128 billion parameters) and DeepSeek V4 Pro (1.6 trillion parameters) were instructed to generate a short example of Dyck-2, the balanced parentheses language with two different types of parentheses.}\label{wrap-fig:1}
\end{figure}

When LLMs generate text, they produce a 
probability distribution over all potential next tokens, from which a new token is sampled. 
%
Recent successes in constrained autoregressive generation mask the next token logits generated by the LLM to ensure that it is only possible for the model to generate valid tokens \citep{deutsch-etal-2019-general, dong2025xgrammar, kuchnik2023validating}. This approach ensures that the models never generate sequences that violate the specified generation constraints. However, the probability of the remainder of the constrained sequence is not incorporated into the estimate of the immediate next token probability. Failing to incorporate future consequences of the immediate next token decision leads to lower quality overall sequence generation \citep{loula2025syntacticsemanticcontrollarge}.

We build on \cite{zhang2023tractablecontrolautoregressivelanguage}, incorporating the future consequences of the next token decision by distilling the LLM into a tractable probabilistic model (TPM), \citep{vergari2019tractable}, in which the future constrained probability mass can be precisely and exactly accounted for. This approach has previously been applied to enforce deterministic finite automata (DFA) constraints on language model generation \cite{zhang2023tractable,zhang2024adaptablelogicalcontrollarge}. DFA constraints can ensure that the LLM generates text that conforms to any specified \textit{regular language}. Regular languages can describe constraints on the inclusion or exclusion of substrings as well as other simple constraints. They cannot enforce the computationally more complex class of \textit{deterministic context-free languages} (DCFLs) \citep{Hopcroft1979IntroductionTA}, which describe languages with recursive structure. Examples of DCFLs range from simple languages such as nested parentheses to complex programming languages like HTML, C, SQL, etc. In order to ensure the satisfaction of an arbitrary DCFL, the appropriate constraining class of automata is deterministic pushdown automata.

This paper documents our method for tractably constraining and steering language models to satisfy DCFLs, which we call \textbf{PASTA-G} (Pushdown-Automata Steering for Tractable Autoregressive Generation). PASTA-G modifies autoregressive language models to exactly incorporate the probability of the entire sequence into each token's probability distribution, improving the generation of complex, recursively defined constraints. In doing so, it enables guaranteed satisfaction of DCFL constraints.

\section{Prior work}
\cite{zhang2023tractablecontrolautoregressivelanguage} describes a method for using a distilled TPM over token sequence probabilities to guide the generation of an autoregressive LLM. Unlike autoregressive LLMs, TPMs can tractably evaluate next token probabilities given complex long-term constraints over future tokens, not just the immediate next token in generation. Constraints (represented as $\alpha$) can be described as a boolean function over sequences. The satisfaction of such a constraint can be described as a weighted model counting problem: the probability of satisfying the boolean constraint $\alpha$ is equal to the sum of the probabilities of all sequences ($x_{1:n}$) which satisfy the constraint.

$$p(\alpha)=\sum_{\mathclap{x_{1:n}\text{ satisfies }\alpha}}p(x_{1:n})$$

In autoregressive generation, where $t-1$ tokens have already been generated, the probability of prior tokens, $x_{<t}$, and the probability of a prospective next token $x_t$ are conditioned upon.

$$p(\alpha\mid x_t, x_{<t})=\sum_{\mathclap{x_{1:n}\text{ satisfies }\alpha}}p(x_{t+1:n}\mid x_t,x_{<t})$$

In the context of this task, a probabilistic model over token sequences is tractable if it is capable of exactly computing such a weighted-model count in polynomial time. If this distilled TPM can estimate the probability of a continuation of the sequence satisfying a logical constraint $\alpha$, then \cite{zhang2023tractablecontrolautoregressivelanguage} demonstrates that an autoregressive LM can be combined with the TPM to produce high quality token sequences guaranteed to satisfy $\alpha$. 
The resulting distribution is:
$$p(x_{t+1} |x_{1:t},\alpha)\propto p_{\text{TPM}}(\alpha|x_{1:t+1})\cdot p_{\text{LM}}(x_{t+1}|x_{1:t})$$

This was further developed in \cite{zhang2024adaptablelogicalcontrollarge}, which showed that this approach can be applied in general to ensure that LM generation satisfies arbitrary deterministic finite automata (DFAs). To show this, they distill the LM to a hidden Markov model (HMM), and demonstrate that the HMM can be used to evaluate the probability of the overall sequence (with a given maximum length) satisfying any DFA. One limitation of this paper is that DFAs can only model regular languages, a class of languages less expressive than most data languages and programming languages \citep{Hopcroft1979IntroductionTA}.

\section{Deterministic pushdown automata}
Pushdown automata can be defined in a variety of ways. Here we only treat stack-empty pushdown automata, a variety that accepts a string only when the automata's stack is empty. These are proven to be equivalent to every other formalization of pushdown automata in \cite{Hopcroft1979IntroductionTA}. In that same text, stack-empty pushdown automata are defined as:
$$A = (Q, \Sigma, \Gamma, q_0, \gamma_0, \delta )$$
where

\begin{itemize}
\item $Q$ is the finite set of states
\item $\Sigma$ is a finite set called the input alphabet
\item $\Gamma$ is a finite set called the stack alphabet
\item $q_0 \in Q$ the start state
\item $\gamma_0 \in \Gamma$ the initial stack symbol
\item $\delta$ is a finite subset of $Q \times (\Sigma \cup \{\epsilon\})\times \Gamma \times Q \times \Gamma^*$  , the transition relation
\end{itemize}

When executing, the pushdown automaton is described as having a configuration. $$C=(q,\gamma),\,q\in Q,\,\gamma\in\Gamma^*$$

Sometimes the stack is described as having a top $v$ and a remainder $W$. $$C=(q,vW), v\in\Gamma, W\in\Gamma^*$$

If the stack is empty, it is written as $$C=(q,\epsilon)$$

Each element of the transition relation associates a state, a stack element, and an input symbol ($\sigma \in \Sigma$) to an output state and a sequence of stack elements. $$(q,\sigma,v,\hat q,V)\in\delta$$

The above rule states that when the pushdown automaton's configuration is ($q$, $vW$) and the next input symbol is $\sigma$, then the automaton's resulting configuration after consuming that symbol will be $(\hat q, VW)$. Importantly, the value $V$ can be a sequence of elements in $\Gamma$ of any length, including length $0$.

The deterministic pushdown automaton (DPDA) requires that if there is an $\epsilon$ transition (a transition requiring no $\sigma$ token as input) available from a configuration, then there can be no $\sigma$ transition defined on that configuration. It also requires that there be at most one $\epsilon$ transition from that configuration. In other words, all $\epsilon$ transitions are required to be processed when in their input configuration.

\section{Approximation of DPDAs using DFAs}
One approach to incorporating the probability of satisfying a DPDA could be to use the prior literature on finite automata satisfaction to approximate a DPDA of finite duration, obviating the need for a bespoke method handling DPDAs. While pushdown automata of finite duration can be approximated by finite state automata, this approximation incurs exponential computational cost. The cache used in \cite{zhang2024adaptablelogicalcontrollarge} is of size $O(m\cdot|Z|\cdot|\Sigma|)$, where $m$ is the number of edges in the DFA, $|Z|$ is the number of hidden states in the HMM, and $|\Sigma|$ is the number of input symbols. In the approximation of a duration-bounded DPDA using a DFA, every state of the DFA represents a configuration of the pushdown automaton ($|Q|\cdot|\Gamma|^H$), where $H$ represents an upper bound on the stack height. The worst case complexity for the number of edges in such a DFA is equal to the number of states squared, leading to a cache size of $O(|Q|^2\cdot|\Gamma|^{2H}\cdot |Z|\cdot |\Sigma|)$. This quickly becomes infeasible to compute as stack height increases, which we explore in our experiments.

\section{Exact inference over deterministic pushdown automata}

There are a variety of tractable probabilistic models that could be used to calculate the probability of satisfying a DPDA. Although PASTA-G follows the prior literature in using an HMM as its TPM, we first describe how exact inference over DPDAs is calculated in a fully factorized probabilistic model to provide intuition for the more complex HMM derivation.

\subsection{Fully factorized model}

In the fully factorized model, the probability of each token is independent, which means the probability of an entire token sequence is the product of the probabilities of each token in the sequence ($p(x_{1:n})=\prod_{i}p(x_i)$). Building on \cite{zhang2024adaptablelogicalcontrollarge}, we need to evaluate the probability that the constraint $\alpha$ is true given all tokens up to and including the current token of generation. For a stack-empty DPDA, the constraint we plan to satisfy is that the stack is empty after generating $n$ tokens ($n$ being a target for the final duration of the entire sequence).
\begin{equation*}
    p(\alpha\mid x_t,x_{<t})=p(\gamma_n=\epsilon\mid x_{t},x_{<t})
\end{equation*}

Using the law of total probability, we can introduce the state of the DPDA here, describing the entire configuration of the DPDA.
\begin{equation*}
    p(\gamma_n=\epsilon\mid x_{t},x_{<t})=\sum_{q_n\in Q}p(q_n,\gamma_n=\epsilon\mid x_t, x_{<t})
\end{equation*}


After consuming $t$ tokens, all future configurations of a DPDA are entirely dependent on the configuration at time $t$. The DPDA cannot "look back" at prior tokens to modify its configuration later on in processing. This Markov property of DPDAs allows the DPDA configuration to substitute for the observed tokens due to the conditional independence $(q_n,\gamma_n)\indep x_{1:t}\mid(q_t,\gamma_t)$. Using this independence, we can substitute the configuration at time $t$ for the tokens in the prior equation.
\begin{equation*}
    p(q_n, \gamma_n=\epsilon\mid x_{t},x_{<t})=p(q_n,\gamma_n=\epsilon\mid q_t, \gamma_t)
\end{equation*}

Assume that $\gamma_t=vW$, where $v\in\Gamma$ is the top of the stack and $W\in\Gamma^*$ is the remainder of the stack. In order for the stack to be reduced to $\epsilon$ in $n-t$ tokens, the top value of the stack, $v$, must be reduced to $\epsilon$ in less than or equal to $n-t$ tokens, and the remainder of the stack $W$ must be reduced to $\epsilon$ in the remaining number of tokens. In enumerating every path of reduction, we must also sum over all potential configurations at each potential cut.\footnote{We provide a derivation of equation 1 in appendix B.1.}
\begin{align}
 p(&q_n,\gamma_n=\epsilon\mid q_t, \gamma_t=vW)=\nonumber
\\&\sum_{u=0}^{n-t}\sum_{q_{t+u} \in Q}
 p(q_{t+u},\gamma_{t+u}=\epsilon\mid q_t,\gamma_t=v) \cdot
 p(q_n,\gamma_n=\epsilon\mid q_{t+u},\gamma_{t+u}=W)
\end{align}

This grounds the probability of reducing an entire stack in the probabilities of reducing individual elements of the stack.  The single stack element component of equation 1 is the probability that a configuration with a single stack element $v$ reduces to $\epsilon$ in $u$ tokens. The value of this depends on whether the configuration $(q_t,\gamma_t)$ has an $\epsilon$ transition or not. If the configuration has an $\epsilon$ transition, then that transition must be applied immediately by the definition of the DPDA.
\begin{align}
(q_t,&\epsilon,v,\hat q_{t},V)\in \delta \implies \nonumber
\\
&p(q_n,\gamma_n=\epsilon\mid q_t,\gamma_t=v) = p(q_n,\gamma_n=\epsilon\mid \hat q_t, \gamma_t=V)
\end{align}

If there is no $\epsilon$ transition defined on the configuration at time $t$, the probability of a single stack element reducing to $\epsilon$ in $n-t$ tokens can be described as the probability that the next token is a valid transition token from the input configuration composed with the probability that the resulting configuration is satisfied in $n-t-1$ tokens. For the purpose of concise notation, we define a function $\Delta$, which takes a configuration and returns all token-state-stack triples into which that configuration can progress.
\begin{equation*}
    (\sigma,q_{t+1},V)\in\Delta(q_t,v) \iff (q_t,\sigma,v,q_{t+1},V)\in \delta
\end{equation*}

Using this function, we can define our single-stack element reduction equation:
\begin{align}
(q_t,&\epsilon,v,\hat q_{t},V)\not\in \delta \implies\nonumber
\\&
p(q_n,\gamma_n=\epsilon\mid q_t,\gamma_t=v) = 
\sum_{\mathclap{(x_{t+1},q_{t+1},\gamma_{t+1})\in \Delta(q_t,v)}} p(x_{t+1})\cdot 
p(q_n,\gamma_n=\epsilon\mid q_{t+1},\gamma_{t+1})
\end{align}

While the last probability statement may condition on a stack $\gamma_{t+1}$ with multiple elements in it, the duration of the remaining sequence is reduced by 1 token. In combination with the multi-element stack equation, this enables the calculation of $p(q_n,\gamma_n=\epsilon\mid q_t,\gamma_t)$ given prior knowledge of $p(q_n,\gamma_n=\epsilon\mid q_{t+u},\gamma_{t+u})$ for all $q\in Q, \gamma_{t+u}\in\Gamma, 1\leq u\leq n-t$. This composition is described in detail in the appendix A.1, under Algorithm 1.

The base case of this relation is when $t=n$, at which point if there are $\epsilon$ transitions defined on the configuration $(q_n,\gamma_n)$, the probability of reducing to $\epsilon$ in $0$ tokens is the probability that the resulting configuration after applying the $\epsilon$ transition reduces to $\epsilon$ in $0$ tokens (this merely restates equation 2). If there are no $\epsilon$ transitions defined on the configuration at token $n$, then the probability that the constraint is satisfied is 1 if and only if the stack is empty.
\begin{align*}
(q_t,&\epsilon,a,\hat q_t, \hat a)\not \in \delta \implies\\
&p(q_n,\gamma_n=\epsilon\mid q_n,\gamma_n)=\mathbbold 1[\gamma_n=\epsilon]
\end{align*}

Due to the expensive nature of recomputing equations 2 and 3 for every configuration and duration, a cache can be constructed beginning at the base case of $p(q_n,\gamma_n=\epsilon\mid q_n,\gamma_n=v)$ and progressing to $p(q_n,\gamma_n\mid q_0,\gamma_0=v)$.  Elements of this cache can be dynamically composed at runtime using equation 1, allowing for the estimation of satisfying a stack of arbitrary height $h$ in $O(|Q|^2\cdot n^2\cdot h)$ floating point multiplications (see appendix A.1). The runtime to construct the cache is $O(n^3\cdot |Q|^2\cdot|\delta|\cdot h)$, where $h$ is the maximum number of stack additions for any transition rule. This can be understood as applying Algorithm 1 to each transition rule in $\delta$ for successively larger values of $n$.

\newpage
\subsection{Hidden Markov Model (HMM)}
\begin{wrapfigure}{r}{3.5cm}
\includegraphics[width=3.5cm]{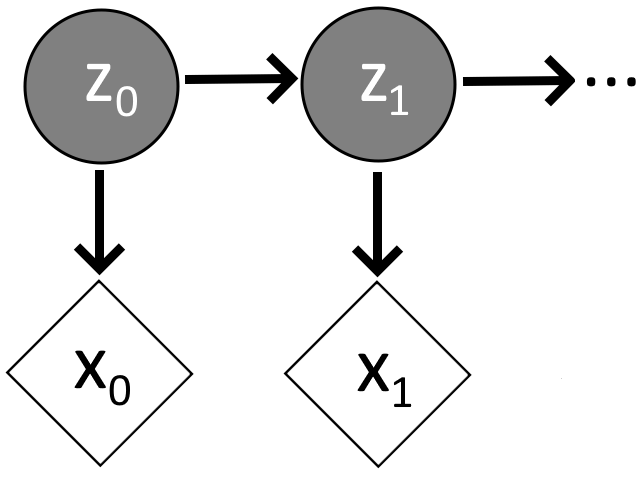}
\caption{HMM}\label{wrap-fig:1}
\end{wrapfigure} 
In the context of autoregressive language generation, a hidden Markov model (HMM) is a probability distribution over sequences of observed tokens. Unlike the fully-factorized example treated in section 5.1, in an HMM tokens are not independent. Rather, tokens are conditionally independent given a hidden state of the model $z_t\in Z$, which may change in value every time a token is emitted. This is the tractable probabilistic model used in \cite{zhang2024adaptablelogicalcontrollarge} to ensure language models satisfy DFA constraints.

Using the same approach as section 5.1, the constraint our TPM must calculate, $\alpha$, represents the stack being empty after generating our target $n$ tokens. This constraint ensures that the final sequence will be of duration $n$ and will be accepted by the deterministic pushdown automaton.
\begin{equation*}
p(\alpha|x_t,x_{<t})=p(\gamma_n=\epsilon|x_t,x_{<t})=\dfrac{p(\gamma_n=\epsilon, x_t,x_{<t})}{p(x_t,x_{<t})}
\end{equation*}

The denominator is easy to compute, so we focus on the numerator. We use the law of total probability to introduce $z_t$, the hidden state of our HMM.
\begin{equation*}   p(\gamma_n=\epsilon,x_t,x_{<t})=\sum_{z_t\in Z}p(\gamma_n=\epsilon|z_t,x_t,x_{<t})\cdot p(z_t,x_t,x_{<t})
\end{equation*}

In generating and parsing sequentially using a DPDA, after parsing the first $t$ tokens, the future configurations of the DPDA depend solely on the configuration of the DPDA immediately after consuming token $t$. As a reminder, the DPDA cannot look back at prior tokens or configurations to determine its state later during processing. This is a Markov property of DPDAs and allows us to write: $p(\gamma_n=\epsilon\mid z_t,x_t,x_{<t})=p(\gamma_n=\epsilon\mid z_t,q_t,\gamma_t)$. Applying that identity, we can write
\begin{equation*}
p(\gamma_n=\epsilon,x_t,x_{<t})=\sum_{z_t\in Z}\boxed{p(\gamma_n=\epsilon|z_t,q_t,\gamma_t)}\cdot p(z_t,x_t,x_{<t})
\end{equation*}

We can incorporate the potential final states $q_n$ as well as potential final HMM hidden states $z_n$ into the equation using the law of total probability:
$$\boxed{p(\gamma_n=\epsilon|z_t,q_t,\gamma_t)}=\sum_{q_n \in Q}
\sum_{z_n\in Z} p(z_n, q_n,\gamma_n=\epsilon|z_t,q_t,\gamma_t)$$

Assume that $\gamma_t=vW$, where $v\in\Gamma$ is a single stack symbol, and $W\in\Gamma^*$ is the remainder of the stack. For $\gamma_n$ to be $\epsilon$ given the stack $\gamma_t=vW$, first the top element of $\gamma_t$ ($v$) must be reduced to $\epsilon$ in fewer than $n-t$ tokens, and then the remainder of the stack ($W$) must be reduced to $\epsilon$ in the remaining tokens. Summing over all possible durations $u$ of reducing $v$ to $\epsilon$, all possible HMM states $z_{t+u}$ after reducing $v$ to $\epsilon$ in $u$ tokens, and all possible DPDA states $q_{t+u}$, we can isolate the probability of reducing the top element of the stack $v$ from the probability of reducing the remainder of the stack $W$.\footnote{We derive equation 4 in detail in appendix B.2.}
\begin{align}
&p(z_n, q_n,\gamma_n=\epsilon\mid z_t,q_t,\gamma_t=vW) =\nonumber 
\\&
\sum_{u=0}^{n-t} \sum_{q_{t+u}\in Q} \sum_{z_{t+u}\in Z}
p( z_{t+u}, q_{t+u},\gamma_{t+u}=\epsilon \mid z_t, q_t,\gamma_t=v)\cdot
p(z_{n},q_n,\gamma_n=\epsilon\mid z_{t+u},q_{t+u},\gamma_{t+u}=W)
\end{align}

The first component of equation 4 is the joint probability that a configuration with a single stack element $v$ reduces to $\epsilon$ in $u$ tokens, and that the HMM's hidden state at time $t+u$ is $z_{t+u}$ given the HMM's hidden state at time $t$. The value of this depends on whether the configuration $(q_t,\gamma_t)$ has an $\epsilon$ transition or not. If the configuration has an $\epsilon$ transition, then that transition must be applied immediately by the definition of the DPDA.
\begin{align}
    (q_{t},& \epsilon,v,\hat q_t,V)\in\delta \implies\nonumber
    \\ 
    & p(z_{t+u},q_{t+u},\gamma_{t+u}=\epsilon\mid z_t, q_t,\gamma_t=v) = p(z_{t+u},q_{t+u},\gamma_{t+u}=\epsilon\mid z_t, \hat q_t,\gamma_t=V)
\end{align}

If there is no $\epsilon$ transition defined on the configuration at time $t$, the probability of a single stack element reducing to $\epsilon$ in $u$ tokens can be described as the probability that the next token is a valid transition token from the input configuration composed with the probability that the resulting configuration is reduced in $n-1$ tokens. To describe this succinctly, we borrow the function $\Delta$ from section 5.1. The function $\Delta$ takes a configuration and returns all token-state-stack triples into which that configuration can progress.
\begin{equation*}
    \Delta(q_t,v)=(\sigma,q_{t+1},V) \iff (q_t,\sigma,v,q_{t+1},V)\in \delta
\end{equation*}

Using this function, we can define our probability of reducing a single stack element when there is no $\epsilon$ transition defined on it.
\begin{alignat}{3}
(q_{t},& \epsilon,v,q_t',V)\not\in\delta \implies
&\nonumber \\
&p(z_{t+u}, q_{t+u}, \gamma_{t+u}=\epsilon\mid z_t, q_t, \gamma_t=v) =\nonumber \\& \sum_{z_{t+1}}p(z_{t+1}|z_t)
\sum_{\mathclap{(x_{t+1},q_{t+1},\gamma_{t+1})\in \Delta(q_t,v)}} p(x_{t+1}|z_{t+1})\cdot
p(z_{t+u}, q_{t+u}, \gamma_{t+u}=\epsilon\mid z_{t+1}, q_{t+1},\gamma_{t+1})
\end{alignat}

While equation 6 may increase the stack size (as with equation 3), it will always decrease the duration of the remaining sequence. The exact value of equation 6's recurrent component can be calculated using equation 4. The base case for equation 6 is when $u=0$. When this is the case, if there are $\epsilon$ transitions defined on the configuration, then the probability of reducing the stack is calculated using equation 5. If there are no $\epsilon$ transitions defined on the configuration, then the probability of reducing the stack to $\epsilon$ is 1 if and only if the stack is already empty.
\begin{align*}
(q_t,&\epsilon,v,\hat q_t, V)\not \in \delta \implies\\
&p(z_t, q_t,\gamma_t=\epsilon\mid z_t,q_t,\gamma_t)=\mathbbold 1[\gamma_t=\epsilon]
\end{align*}

The relationships defined above monotonically reduce the sequence duration required to compute their probabilities. This allows us to compute the value of equation 4 for sequences of duration $n-t$ knowing only the values of equations 4, 5, and 6 for sequence durations less than $n-t$. Using this fact, we can iteratively construct from base cases a cache of values representing equations 5 and 6 and compose those values to evaluate equation 4. The cache for equations 5 and 6 occupies $O(|Z|^2\cdot|Q|^2\cdot |\Gamma|\cdot n)$ space. The composition of these values into equation 4 for a stack height of $h$ runs in $O(n^2\cdot |Q|^2\cdot |Z|^2\cdot h)$ floating point multiplications using Algorithm 2 specified in appendix A.2. By applying Algorithm 2 for every transition rule in $\delta$ and every duration from $0$ to $n$, the cache can be constructed in $O(n^3\cdot |Q|^2\cdot  |Z|^2\cdot h\cdot |\delta|)$ time.
\newpage
\section{Experiments}
\begin{wrapfigure}{r}{5cm}
\includegraphics[width=5cm]{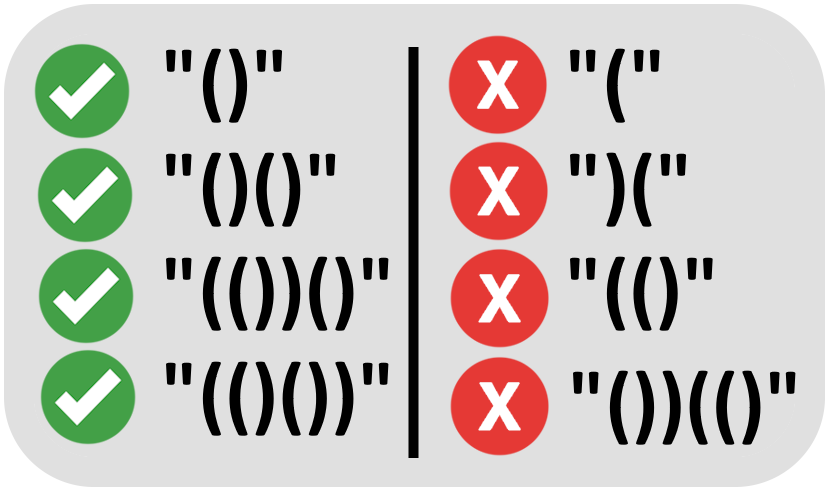}
\caption{Balanced parentheses (Dyck-1)}\label{wrap-fig:1}
\end{wrapfigure} 

To empirically validate PASTA-G, we implemented it in Python and compared its estimation of $p(\alpha\mid x_t,x_{<t})$ with the brute-force weighted-model count estimation of $p(\alpha\mid x_t,x_{<t})$. We applied these tests on the prototypical balanced-parentheses languages, comparing runtime and memory usage with Ctrl-G.

Balanced parentheses languages (also called Dyck languages) are those languages that are only satisfied when their input is a properly nested sequence of parentheses. An illustration of balanced parentheses (Dyck-1) strings can be found in Figure 3. Dyck-1 is context-free, requires hierarchical and recursive parsing, and cannot be represented by a finite-state automaton \citep{Hopcroft1979IntroductionTA}, making it a good candidate for comparing PASTA-G to Ctrl-G. 


One way to evaluate the empirical accuracy of PASTA-G is by treating it as a weighted model counting problem. We can evaluate the true probability of constraint satisfaction by looking at all potential strings and evaluating the sum of the probabilities of those strings that satisfy the constraint. From an autoregressive context, we must condition on the already generated content and sum over potential string completions.
\begin{align*}
p(\alpha\mid x_t, x_{<t})&=\sum_{\mathclap{x_{1:n} \text{ satisfies } \alpha}}p(x_{t+1:n}\mid x_t, x_{<t})
\end{align*}

We test PASTA-G's expected probability distribution over possible next tokens $x_t$ using all potential prefixes for expected sequence lengths between 1 and 10 across a collection of randomly initialized two-state HMMs. For all next tokens ($x_t\in\{``(",``)",``\text{<EOS>}"\}$), prefixes ($x_{1:t-1}$), sequence lengths ($n$), and HMM initializations, we see identical probabilities between our evaluation and the weighted-model counting evaluation of $p(\alpha\mid x_t, x_{1:t-1})$.

To evaluate the runtime complexity, we benchmark both the cache size and inference time as they vary with the maximum sequence length in estimating the probability of Dyck-1. Because Dyck-1 can pop at most one stack element per token, we can upper-bound the stack-height as $H=\lceil n/2 \rceil$, where $n$ represents the upper-bound on how many tokens can be generated. Using this upper-bound, we can approximate a length-bounded Dyck pushdown automaton by creating a finite set of states $(q,\gamma),\forall q\in Q,\gamma\in\Gamma^h,0\leq h\leq H$. This set of states can be used as input to Ctrl-G to simulate the behavior of our pushdown automaton probability estimation. We find that, as expected, Ctrl-G has exponential cache-space and inference-time complexity with respect to sequence length, whereas PASTA-G has linear space-complexity and quadratic inference-time complexity with respect to sequence length, as shown in Figure 4.

\begin{figure}[h]
\includegraphics[width=15cm]{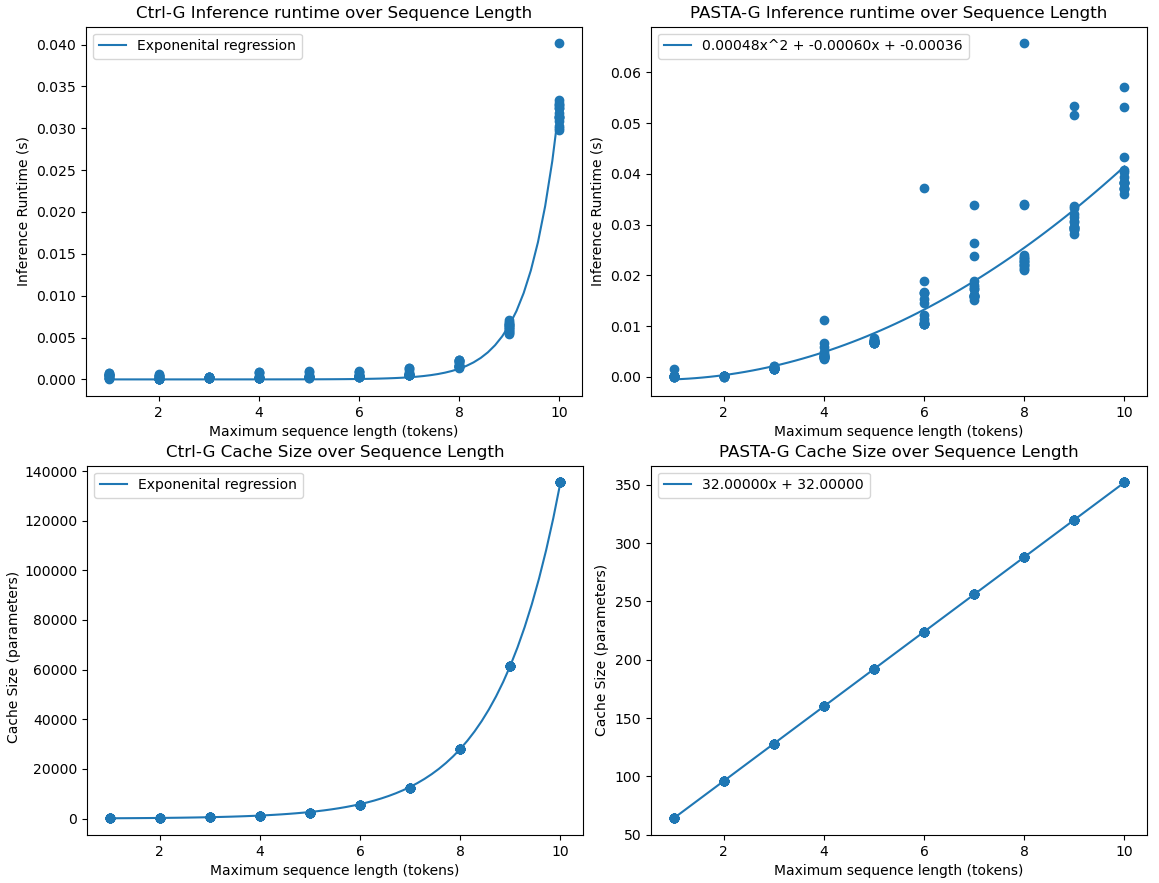}
\centering
\caption{\textbf{Runtime and cache size of Ctrl-G and PASTA-G for Dyck-1.} As sequence length increases, Ctrl-G's cache size and inference time grow exponentially, while PASTA-G's cache size grows linearly and its inference time grows quadratically.}\label{wrap-fig:1}
\centering
\end{figure} 

\section{Related Work}
In \cite{loula2025syntacticsemanticcontrollarge},
the incorporation of total sequence probability into the autoregressive next token decision for context-free grammar satisfaction is achieved using sequential Monte-Carlo. 
This approach demonstrates impressive results across a variety of languages. 
It does not, however, use the tractable marginalization enabled by a TPM, which implies that the method would rely on a large number of particles to converge.
It does show that logit-based steering can measurably improve generation quality for complex real-world code and data constrained generation tasks.


\cite{baiget2026explainfuzzexplainableconstraintconditionedtest}
take a similar approach to integrating TPMs and context-free grammars, though without using LLMs. 
They approach this task by compiling the grammar into a tractable probabilistic circuit (PC) \citep{darwiche2003differential,vergari2019tractable,choi2020pc} that can be trained on observed examples of the grammar. 
PCs generalize HMMs and other TPMs and have been extensively used for tractable neuro-symbolic computation \citep{ahmed2022}.
PASTA-G uses a preexisting TPM (an HMM) and composes its parameters dynamically to tractably estimate the probability of each element in a sequence autoregressively.
Ideally, we could use the same compilation strategy of \cite{baiget2026explainfuzzexplainableconstraintconditionedtest}, which incurs cubic complexity, to obtain a PC that encodes a context-free grammar and tractably multiply it with an HMM to retrieve our algorithm \citep{vergari2021compositional}.
This would, however, require additional memory and computational overhead for the multiplication. 
Two further advantages of our method over that used in 
\cite{baiget2026explainfuzzexplainableconstraintconditionedtest}
are that PASTA-G does not require re-distillation of the tractable probabilistic model when the grammar changes, and PASTA-G's HMM has no upper-bound on its parameter count for grammars generated to fixed sequence durations.

\section{Future areas of research}
\subsection{Tokenization}
One limitation in applying PASTA-G to large language models is the problem of misalignment between the LLM tokenizer and the tokenization assumed in the DPDA. The first issue with such misalignment is that the LLM tokenizer may split DPDA tokens, breaking the determinism required by this approach. Another issue is that the DPDA tokens may split canonical sequence tokenizations for the LLM. For example, most deterministic context-free grammar representations of English word sequences assume spaces are separate tokens from words, whereas most byte-pair-encoding tokenizers used in LLMs merge words and spaces. These tokenization misalignments will need to be addressed to properly apply PASTA-G to more diverse languages.

\subsection{Other future areas of research}
Any language that can be parsed deterministically using a definite number of look-ahead symbols is a deterministic context-free language and can be parsed using a DPDA \citep{KNUTH1965607}. Automatically translating DCFLs more complex than those treated in this paper into DPDAs could be a fruitful area of future research. Doing so would enable the effortless application of PASTA-G to arbitrary DCFLs.

The extension of this work to other formal automata is an open area of future research. Applying PASTA-G's approach to nondeterministic pushdown automata would generate more powerful languages with exact satisfaction guarantees. Similarly, creating a method similar to PASTA-G for more powerful automata such as the nested stack automaton would enable its application to context-sensitive languages. Most natural languages are "mildly context-sensitive," \citep{bb1d1d0f-0e14-384a-9afc-6e7e295f8353, 10.5555/645667.665050}, and some programming languages are as well (e.g., Python, \cite{10.1145/2997364.2997370}). Extending exact probabilistic inference to constrained generation of mildly context-sensitive languages could improve autoregressive code and language generation.
Finally, our approach could be hybridized with that in \citet{baiget2026explainfuzzexplainableconstraintconditionedtest} by embedding an HMM as a probabilistic circuit layer in an LLM \citep{grivas2026fast} that compiles a context-free grammar.

\bibliographystyle{plainnat}
\bibliography{main}

\newpage

\appendix
\section{Dynamic Programming Algorithms}
\subsection{Algorithm 1}
\begin{algorithm}[H]
\caption{Evaluate equation 1 to compute $p(q_n,\gamma_n=\epsilon\mid q_t, \gamma_t)$}
\label{alg:eq1-eval}
\begin{algorithmic}[1]

\Require Cache of probabilities $C[u, q_u,q_1,\gamma_1]=p(q_u,\gamma_u=\epsilon\mid q_1,\gamma_1)$

\Require Configuration $(q_t, \gamma_t)$ where $q_t \in Q$, $\gamma_t \in \Gamma^h$ 

\State $\textit{acc} \gets \mathbf{0}$ \Comment{$|Q| \times (n-t+1)$ matrix}
\Statex

\ForAll{$q \in Q$, $q' \in Q$}
\Comment{Initialize accumulator using bottom-of-stack $\gamma_t[h-1]$}
    \For{$d \gets 0$ \textbf{to} $n-t$}
      \State $\textit{acc}[q',d] \mathrel{+}= C[d,q',q,\gamma_t[h-1]]$
    \EndFor
\EndFor
\Statex

\For{$s \gets h-2$ \textbf{down to} $0$}
  \State $\textit{acc}' \gets \mathbf{0}$ \Comment{$|Q| \times (n-t+1)$ matrix}
  \For{$d \gets 0$ \textbf{to} $n-t$}
  \Comment{Duration $d$}
    \For{$u \gets 0$ \textbf{to} $d$}
      \Comment{Split index $u$}
      \ForAll{$q \in Q$, $q' \in Q$}
          
          \Comment{Restatement of equation 1}
          \State $\textit{acc}'[q',d] \mathrel{+}= C[u, q',q,\gamma_t[s]) \cdot \textit{acc}[q,d-u]$
      \EndFor
    \EndFor
  \EndFor
  \State $\textit{acc} \gets \textit{acc}'$
\EndFor
\Statex

\State \Return $\textit{acc}[q_t,\, n-t]$
\end{algorithmic}
\end{algorithm}
\newpage

\subsection{Algorithm 2}
\begin{algorithm}[H]
\caption{Evaluate equation 4 to compute $p(z_n, q_n,\gamma_n=\epsilon\mid z_t,q_t, \gamma_t)$}
\label{alg:eq1-eval}
\begin{algorithmic}[1]

\Require Cache of probabilities $C[u, z_u,q_u,z_1,q_1,\gamma_1]=p(z_u, q_u,\gamma_u=\epsilon\mid z_1,q_1,\gamma_1)$

\Require Configuration $(q_t, \gamma_t)$ where $q_t \in Q$, $\gamma_t \in \Gamma^h$ 

\State $\textit{acc} \gets \mathbf{0}$ \Comment{$|Q| \times (n-t+1) \times |Z|$ matrix}
\Statex

\ForAll{$q \in Q$, $q' \in Q$}
\Comment{Initialize accumulator using bottom-of-stack $\gamma_t[h-1]$}
    \ForAll{$z\in Z,z' \in Z$}
        \For{$d \gets 0$ \textbf{to} $n-t$}
          \State $\textit{acc}[q',d,z'] \mathrel{+}= C[d,z',q',z,q,\gamma_t[h-1]]$
        \EndFor
  \EndFor
\EndFor
\Statex

\For{$s \gets h-2$ \textbf{down to} $0$}
  \State $\textit{acc}' \gets \mathbf{0}$ \Comment{$|Q| \times (n-t+1) \times |Z|$ matrix}
  \For{$d \gets 0$ \textbf{to} $n-t$}
  \Comment{Duration $d$}
    \For{$u \gets 0$ \textbf{to} $d$}
      \Comment{Split index $u$}
      \ForAll{$q \in Q$, $q' \in Q$}
        \ForAll{$z \in Z$, $z' \in Z$}
          
          \Comment{Restatement of equation 4}
          \State $\textit{acc}'[q',d, z'] \mathrel{+}= C[u, z', q',z,q,\gamma_t[s]) \cdot \textit{acc}[q,d-u,z]$
          \EndFor
      \EndFor
    \EndFor
  \EndFor
  \State $\textit{acc} \gets \textit{acc}'$
\EndFor
\Statex

\State \Return $\textit{acc}[q_t,\, n-t, z_t]$
\end{algorithmic}
\end{algorithm}

\newpage
\section{Derivations of stack decomposability equations}
\subsection{Derivation of equation 1}

To derive equation 1, we must ground the expression below in probability expressions conditioned upon single stack elements, rather than the entire stack $\gamma_t$.

$$p(q_n, \gamma_n=\epsilon\mid q_t, \gamma_t)$$

For the entire stack $vW$ to reduce to $\epsilon$ in $n-t$ tokens, at some point in the reduction the stack must be equal to $W$. Applying any transition consumes the top element of the stack and pushes some number of stack elements back onto the stack. The minimum number of elements that can be pushed onto the stack is $0$, in which case the stack after applying the transition is $1$ element smaller than the stack prior. No matter how many elements are pushed, each of those elements must be popped one transition at a time for the entire stack to reduce to $\epsilon$. This iterative popping necessarily reduces the stack $vW$ to $W$ at some point in the reduction to $\epsilon$, because the DPDA cannot pop the top element of $W$ without the stack being equal to $W$.

By the law of total probability, we can introduce a cut in the derivation, supposing a number of tokens $u$ and state $q_{t+u}$ which correspond to reducing the stack to $W$ in $u$ tokens without having reduced the stack to $W$ prior to consuming $u$ tokens.
\begin{align*}
     p(q_n,\gamma_n=\epsilon\mid q_t, \gamma_t=vW)=
 \sum_{u=0}^{n-t}\sum_{q_{t+u} \in Q} &
 p(q_{t+u},\gamma_{t+u}=W, W\not\in\gamma_{t:t+u-1}, q_n,\gamma_n=\epsilon\mid q_t, \gamma_t=vW)
\end{align*}

By Bayes theorem we can decompose the above equation.

\begin{align*}
     p(q_n,\gamma_n=\epsilon\mid q_t, \gamma_t=vW)=
 \vphantom{\sum}\smash{\sum_{u=0}^{n-t}\sum_{q_{t+u} \in Q}}\,\, &
 p(q_n,\gamma_n=\epsilon\mid q_t, \gamma_t=vW, q_{t+u},\gamma_{t+u}=W, W\not\in\gamma_{t:t+u-1})\cdot
 \\
&
p(q_{t+u},\gamma_{t+u}=W, W\not\in\gamma_{t:t+u-1}\mid q_t,\gamma_t=vW)
\end{align*}

By the Markovian conditional independence $(q_{t},\gamma_t)\indep(q_{t+k},\gamma_{t+k})\mid(q_{t+l},\gamma_{q+l}),\forall k, l\in \mathbb N, l<k$, future DPDA configurations are isolated from prior configurations given knowledge of an intermediate configuration.

\begin{align*}
     p(q_n,\gamma_n=\epsilon\mid q_t, \gamma_t=vW)=
 \smash{\sum_{u=0}^{n-t}\sum_{q_{t+u} \in Q}}\,\, &
 p(q_n,\gamma_n=\epsilon\mid q_{t+u},\gamma_{t+u}=W)\cdot
 \\
&
\boxed{p(q_{t+u},\gamma_{t+u}=W, W\not\in\gamma_{t:t+u-1}\mid q_t,\gamma_t=vW)}
\end{align*}\\
The boxed probability expression can be rewritten as a weighted model count over all token sequences that satisfy the automata constraints described in the expression. We define $F$ as the set of all token strings which reduce the configuration $(q_t, vW)$ to $(q_{t+u}, W)$ in $u$ tokens without reducing the stack to $W$ in fewer than $u$ tokens: $$F=\{x_{t+1:t+u}:(q_t,\gamma_t=vW)\overset {x_{t+1:t+u}}\longrightarrow (q_{t+u}, \gamma_{t+u}=W) \land W\not\in\gamma_{t:t+u-1}\}$$\\
Using this expression of our constraint, we write the weighted model count as:
$$p(q_{t+u},\gamma_{t+u}=W,W\not\in\gamma_{t:t+u-1}\mid q_t,\gamma_t=vW)=\sum_{\mathclap{x_{t+1:t+u}}}p(x_{t+1:t+u})\cdot\mathbbold{1}[x_{t+1:t+u}\in F]$$

The DPDA can only follow transition rules which correspond to elements on the top of the stack during its derivation. Our constraint requires that at no point in derivation prior to $t+1$ is the stack equal to $W$, meaning the top element of $W$ is never at the top of the stack prior to the final configuration $(q_{t+u},\gamma_{t+u})$. These two facts demonstrate the independence of $(q_t,vW)\overset{x_{t+1:t+u}}\longrightarrow(q_{t+u},W)$ from $W$ when $W\not\in\gamma_{t:t+u-1}$. If the top element of $W$ never affects the transitions taken, then the reduction of the stack to $W$ can be calculated in the absence of $W$. See appendix C for a proof of this equivalence.

$$G=\{x_{t+1:t+u}:(q_t,\gamma_t=v)\overset{x_{t+1:t+u}}\longrightarrow(q_{t+u},\gamma_{t+u}=\epsilon)\}$$
$$F=G$$

Using this logical equivalence, we can rewrite our weighted model count. $$p(q_{t+u},\gamma_{t+u}=W,W\not\in\gamma_{t:t+u-1}\mid q_t,\gamma_t=vW)=\sum_{\mathclap{x_{t+1:t+u}}}p(x_{t+1:t+u})\cdot\mathbbold{1}[x_{t+1:t+u}\in G]$$
\\
The weighted model count can be converted back into a probability statement about the contents of the DPDA configuration.
$$p(q_{t+u},\gamma_{t+u}=W,W\not\in\gamma_{t:t+u-1}\mid q_t,\gamma_t=vW)=p(q_{t+u},\gamma_{t+u}=\epsilon\mid q_t,\gamma_t=v)$$
\\
This equivalence allows us to rewrite our reduction, leaving us with equation 1. \begin{align*}
     p(q_n,\gamma_n=\epsilon\mid q_t, \gamma_t=vW)=
 \sum_{u=0}^{n-t}\sum_{q_{t+u} \in Q} &
 p(q_n,\gamma_n=\epsilon\mid q_{t+u},\gamma_{t+u}=W)\cdot
p(q_{t+u},\gamma_{t+u}=\epsilon\mid q_t,\gamma_t=v)
\end{align*}
\newpage
\subsection{Derivation of equation 4}
Similarly to appendix B.1, to derive equation 4 we must ground the probability expression below in probability expressions conditioned on single stack elements, rather than the entire stack $\gamma_t$.
$$p(z_n,q_n,\gamma_n=\epsilon\mid z_t, q_t, \gamma_t)$$

By the law of total probability, we can introduce a cut at token $u$ into our derivation. For any derivation that pops all elements of the stack $vW$, at some point $u$ tokens in the future the stack must be equal to the remainder of the stack $W$ because the stack can pop at most one element per transition. At that point where the stack is equal to $W$, there will be an intermediate HMM hidden state $z_{t+u}$ and an intermediate DPDA state $q_{t+u}$. We sum over all possible values of both of those variables at this intermediate state in the derivation.

\begin{align*}
p&(z_n, q_n,\gamma_n=\epsilon\mid z_t,q_t,\gamma_t=vW) = \\
&\sum_{u=0}^{n-t}\sum_{q_{t+u}\in Q}\sum_{z_{t+u}}\,p(z_n,q_n,\gamma_n=\epsilon,
z_{t+u},q_{t+u},\gamma_{t+u}=W,
W\not\in\gamma_{t:t+u-1}\mid z_t,q_t,\gamma_t=vW)
\end{align*}

By Bayes' theorem, we can separate the parts of the above expression which don't correspond to the state at token $n$.

\begin{alignat*}{10}
p&(z_n, q_n,\gamma_n=\epsilon\mid z_t,q_t,\gamma_t=vW) = 
\\&
\vphantom{\sum^{1}}\smash{\sum_{u=0}^{n-t}\sum_{q_{t+u}\in Q}\sum_{z_{t+u}}}\,\,
p(z_n,q_n,\gamma_n=\epsilon\mid
z_{t+u},q_{t+u},\gamma_{t+u}=W,
W\not\in\gamma_{t:t+u-1},
z_t,q_t,\gamma_t=vW)\cdot
\\&\hspace{2.4cm}
\mathrlap{p(z_{t+u},q_{t+u},\gamma_{t+u}=W,W\not\in\gamma_{t:t+u-1}\mid z_t,q_t,\gamma_t=vW)}
&&&&
\end{alignat*}
By the Markov independencies $(q_t,\gamma_t)\indep(q_{t+k},\gamma_{t+k})\mid(q_{t+l},\gamma_{t+l})$ and $z_t\indep z_{t+k}\mid z_{t+l}$ for all $k,l\in\mathbb N, l < k$, the state of the HMM and configuration of the DPDA at token $n$ can be isolated from the state and configuration at token $t$ given knowledge of the intermediate state and configuration at token $t+u$.
\begin{align*}
p&(z_n, q_n,\gamma_n=\epsilon\mid z_t,q_t,\gamma_t=vW) = \\&\vphantom{\sum^{1}}\smash{\sum_{u=0}^{n-t}\sum_{q_{t+u}\in Q}\sum_{z_{t+u}}}\,
p(z_n,q_n,\gamma_n=\epsilon\mid
z_{t+u},q_{t+u},\gamma_{t+u}=W)\cdot
\\
&\hspace{2.28cm}\boxed{p(z_{t+u},q_{t+u},\gamma_{t+u}=W,W\not\in\gamma_{t:t+u-1}\mid z_t,q_t,\gamma_t=vW)}
\end{align*}\\
The boxed probability expression can be rewritten as a weighted model count over all sequences that satisfy the automata constraints described in the expression. As in appendix B.1, we define the set of all tokens sequences which satisfy the automata constraint as the set $F$.
$$F=\{x_{t+1:t+u}:(q_t,\gamma_t=vW)\overset {x_{t+1:t+u}}\longrightarrow (q_{t+u}, \gamma_{t+u}=W) \land W\not\in\gamma_{t:t+u-1}\}$$

\begin{align*}
p&(z_{t+u},q_{t+u},\gamma_{t+u}=W,W\not\in\gamma_{t:t+u-1}\mid z_t,q_t,\gamma_t=vW)=\sum_{\mathclap{x_{t+1:t+u}}}\,p(z_{t+u}, x_{t+1:t+u}\mid z_t)\cdot\mathbbold{1}[x_{t+1:t+u}\in F]\end{align*}
\newpage
Using the proof specified in appendix C, this logical equivalence can be rewritten:

$$G=\{x_{t+1:t+u}:(q_t,\gamma_t=v)\overset{x_{t+1:t+u}}\longrightarrow(q_{t+u},\gamma_{t+u}=\epsilon)\}$$
$$F=G$$
\\
This logical equivalence allows us to rewrite our weighted model count as 
\begin{align*}
    p(z_{t+u},q_{t+u},\gamma_{t+u}=W,W\not\in\gamma_{t:t+u-1}\mid z_{t},q_t,\gamma_t=vW)=\sum_{\mathclap{x_{t+1:t+u}}}&\,p(z_{t+u},x_{t+1:t+u}\mid z_t)\cdot\mathbbold{1}[x_{t+1:t+u}\in G]
\end{align*}
\\
The weighted model count can be converted back into a probability statement about the contents of the DPDA configuration.
$$p(z_{t+u},q_{t+u},\gamma_{t+u}=W,W\not\in\gamma_{t:t+u-1}\mid z_t, q_t,\gamma_t=vW)=p(z_{t+u},q_{t+u},\gamma_{t+u}=\epsilon\mid z_t,q_t,\gamma_t=v)$$
\\
Rewriting our reduction using the above equivalence, we get equation 4. \begin{align*}
     p(z_n,q_n,\gamma_n=\epsilon\mid q_t, \gamma_t=vW)=
 \sum_{u=0}^{n-t}\sum_{q_{t+u} \in Q} &
 p(z_n,q_n,\gamma_n=\epsilon\mid z_{t+u},q_{t+u},\gamma_{t+u}=W)\cdot
 \\
&
p(z_{t+u},q_{t+u},\gamma_{t+u}=\epsilon\mid z_t, q_t,\gamma_t=v)
\end{align*}

\newpage
\section{Proof that the stack remainder is irrelevant to reducing the top stack element in exactly $u$ tokens}

Here we prove that the set of token sequences that reduce a DPDA configuration $C_t=(q_t,vW)$ to $C_{t+u}=(q_{t+u},W)$ in exactly $u$ tokens without reducing the stack to $W$ in fewer than $u$ tokens are equivalent to the set of token sequences that reduce the configuration $C_t=(q_t,v)$ to $C_{t+u}=(q_{t+u}, \epsilon)$ in exactly $u$ tokens.

We prove this by demonstrating that at every intermediate configuration between $t$ and $t+u$, $C_i=(q_i,\gamma_i)$, the set of available transitions to the DPDA are identical under these two constraints. This means that the set of tokens accepted at every intermediate configuration are identical, and consuming any of those tokens corresponds to an identical resultant configuration between both of the above conditions (modulo the bottom of the stack, $W$ or $\epsilon$). By that fact, the set of sequences that transition a DPDA $A$ between configurations $(q_{t},vW)\rightarrow (q_{t+u},W)\land W\not\in\gamma_{t:t+u-1}$ is the same as the set of sequences that transition $A$ between configurations $(q_t,v)\rightarrow(q_{t+u},\epsilon)$.

We consider this over the intersection of two sets of two cases. The first set of cases conditions on the structure of the stack. Either there is one stack element, $y$, above $W$/$\epsilon$, or there are multiple elements above $W$/$\epsilon$, which we represent as the list $Y$. The second set of cases is whether the transition being made is the final transition moving into the configuration $C_{t+u}$, or whether more than one transition will be made before entering the final configuration. Across all combinations of these cases, the available sets of transitions to the DPDA are identical.

There may be constraints on the available transitions in a particular configuration that are unique to the DPDA $A$ being analyzed, such as the fact that certain configurations may be impossible to reduce in the required token limit. Under both of the cases, $W$ and $\epsilon$, studied here, those constraints apply as extra filters on the available transitions (and equivalently, the available token sequences). These extra filters all apply to both constraints evaluated, and so have no means of affecting their equivalence.

\subsection{Case 1: The intermediate configuration is in the form $(q_i,yW)$ or $(q_i,y)$ and this is the final transition.}

Under the constraint $(q_t,vW)\overset{x_{t+1:t+u}}\longrightarrow (q_{t+u},W) \land W\not\in\gamma_{t:t+u-1}$, if the configuration is in the form $(q_i, yW)$ and this is the final transition, the DPDA must use the final transition to reduce to $W$ to ensure the stack is $W$ after processing all transitions. The set of transitions available to the DPDA are in the form $(q_i, \cdot,y,q_{t+u},\epsilon)$.

Under the constraint $(q_t,v)\overset{x_{t+1:t+u}}\longrightarrow(q_{t+u},\epsilon)$, if the configuration is in the form $(q_i,y)$ and this is the final transition, for the same reason as in the prior constraint, the DPDA must use the final transition to reduce to $\epsilon$ to satisfy the constraint. The set of transitions available to the DPDA are in the form $(q_i,\cdot,y,q_{t+u},\epsilon)$.

\checkmark The set of transitions available under these two constraints are identical in Case 1.

\subsection{Case 2: The intermediate configuration is in the form $(q_i,Y W)$ or $(q_i,Y)$ and this is the final transition}

Under the constraint $(q_t,vW)\overset{x_{t+1:t+u}}\longrightarrow (q_{t+u},W) \land W\not\in\gamma_{t:t+u-1}$, the list of stack elements $Y$ cannot be reduced in a single transition, because at most one element can be popped from the stack in a single transition, and $Y$ has at least two elements that must be popped for the stack to become $W$. There are no valid transitions in this case.

Under the $(q_t,v)\overset{x_{t+1:t+u}}\longrightarrow (q_{t+u},\epsilon)$ constraint, the list of stack elements $Y$ cannot be reduced in a single transition for the same reason as above. There are no valid transitions in this case either.

\checkmark The set of available transitions under these two constraints are identical in Case 2.

\subsection{Case 3: The intermediate configuration is in the form $(q_i,yW)$ or $(q_i,y)$ and this is not the final transition}

Under the constraint $(q_t,vW)\overset{x_{t+1:t+u}}\longrightarrow (q_{t+u},W) \land W\not\in\gamma_{t:t+u-1}$, pop operations are not allowed in this case, because that would result in a stack $W$ before the final transition, violating the second condition of the constraint. For that reason, the only transition rules allowed in this case are in the form $(q_i,\cdot,y,\hat q,\hat Y), \hat q\in Q, \hat Y\in\Gamma^{\geq1}$.

Under the $(q_t,v)\overset{x_{t+1:t+u}}\longrightarrow (q_{t+u},\epsilon)$ constraint, in this case pop operations are also not allowed, but for a slightly different reason. The transition relation on DPDAs, as described in \cite{Hopcroft1979IntroductionTA}, is a finite subset of $Q \times (\Sigma\cup\{\epsilon\})\times\Gamma\times Q\times\Gamma^*$. The third component of each transition relation is the stack element input required by the transition, specified as an element of $\Gamma$, the set of all stack symbols. Transitions therefore cannot take as their stack element input the empty stack $\epsilon$, as $\epsilon$ is not a stack symbol. This means that after performing a pop operation, no further transitions are possible. However, that would mean that this was the final transition in the production of $(q_{t+u},\epsilon)$, violating this case's constraint that this is not the final transition. Performing a pop operation forces us to violate the constraint of this case, and so is not allowed. Due to this, the set of available transitions in this case and under this constraint are those transition rules in the form $(q_i,\cdot,y,\hat q, \hat Y),\hat q\in Q, \hat Y\in\Gamma^{\geq 1}$

\checkmark The set of available transitions under these two constraints are identical in Case 3.

\subsection{Case 4: The intermediate configuration is in the form $(q_i,Y W)$ or $(q_i,Y)$ and this is not the final transition}

Under the $(q_t,vW)\overset{x_{t+1:t+u}}\longrightarrow (q_{t+u},W) \land W\not\in\gamma_{t:t+u-1}$ constraint, there are no additional constraints on the set of available transitions in this case. Pop, preserve, and push transitions are all valid in this state, because none of them necessarily preclude the possibility of transitioning to the specified final configuration in future transitions.

Under the $(q_t,v)\overset{x_{t+1:t+u}}\longrightarrow (q_{t+u},\epsilon)$ constraint, no additional constraints are put on the set of available transitions in this case for the same reason as above. 

\checkmark The set of available transitions under these two constraints are identical in Case 4.

\subsection{Summary}

In all cases during their derivation, the set of available transitions is identical between both $(q_t,vW)\overset{x_{t+1:t+u}}\longrightarrow (q_{t+u},W) \land W\not\in\gamma_{t:t+u-1}$ and $(q_t,v)\overset{x_{t+1:t+u}}\longrightarrow (q_{t+u},\epsilon)$. This means that the set of token sequences that reduce $(q_t,vW)$ to $(q_{t+u},W)$ without reducing to $W$ before $t+u$ tokens is identical to the set of token sequences which reduce $(q_t,v)$ to $(q_{t+u},\epsilon)$. As sets of token sequences, $$\{x_{t+1:t+u}:(q_t,vW)\overset{x_{t+1:t+u}}\longrightarrow (q_{t+u},W) \land W\not\in\gamma_{t:t+u-1}\}=\{x_{t+1:t+u}:(q_t,v)\overset{x_{t+1:t+u}}\longrightarrow(q_{t+u},\epsilon)\}$$ 

\end{document}